%% file: main.tex
\documentclass[10pt,twocolumn,letterpaper]{article}

\usepackage[pagenumbers]{cvpr} 

\definecolor{cvprblue}{rgb}{0.21,0.49,0.74}
\usepackage[pagebackref,breaklinks,colorlinks,allcolors=cvprblue]{hyperref}
\usepackage{multirow}
\usepackage{makecell}
\usepackage{tcolorbox}
\usepackage{amssymb}
\usepackage{svg}
\usepackage{xspace}

\newcommand{\ourmodel}{SemLayer\xspace}

\title{\ourmodel: Semantic-aware Generative Segmentation and Layer Construction for Abstract Icons}

\author{%
  Haiyang Xu$^{1,2}$ \qquad Ronghuan Wu$^{3}$ \qquad Li-Yi Wei$^{2}$ \qquad Nanxuan Zhao$^{2}$ \\
  Chenxi Liu$^{4}$ \qquad Cuong Nguyen$^{2}$ \quad Zhuowen Tu$^{1}$ \qquad Zhaowen Wang$^{2}$ \\
  $^1$UC San Diego \quad $^2$Adobe Research \quad $^3$City University of Hong Kong \quad $^4$University of Toronto
}

\begin{document}

\maketitle
\input{sec/0_abstract}    
\input{sec/1_intro}
\input{sec/2_related}
\input{sec/3_method}

\input{sec/4_dataset}

\input{sec/5_experiment}
\input{sec/6_conclusion}

{
    \small
    \bibliographystyle{ieeenat_fullname}
    \bibliography{main}
}

\input{sec/X_suppl}

\end{document}

%% file: sec/0_abstract.tex
\begin{abstract}

\vspace{-12pt}
Graphic icons are a cornerstone of modern design workflows, yet they are often distributed as flattened single-path or compound-path graphics, where the original semantic layering is lost. This absence of semantic decomposition hinders downstream tasks such as editing, restyling, and animation. We formalize this problem as semantic layer construction for flattened vector art and introduce \ourmodel, a visual generation empowered pipeline that restores editable layered structures. Given an abstract icon, \ourmodel first generates a chromatically differentiated representation in which distinct semantic components become visually separable. To recover the complete geometry of each part, including occluded regions, we then perform a semantic completion step that reconstructs coherent object-level shapes. Finally, the recovered parts are assembled into a layered vector representation with inferred occlusion relationships. Extensive qualitative comparisons and quantitative evaluations demonstrate the effectiveness of \ourmodel, enabling editing workflows previously inapplicable to flattened vector graphics and establishing semantic layer reconstruction as a practical and valuable task. 
Project page: \url{https://xxuhaiyang.github.io/SemLayer/}

\end{abstract}

%% file: sec/1_intro.tex
\vspace{-18pt}
\section{Introduction}
\vspace{-6pt}
\label{sec:intro}

\begin{figure}[th]
    \centering
    \includegraphics[width=0.95\linewidth]{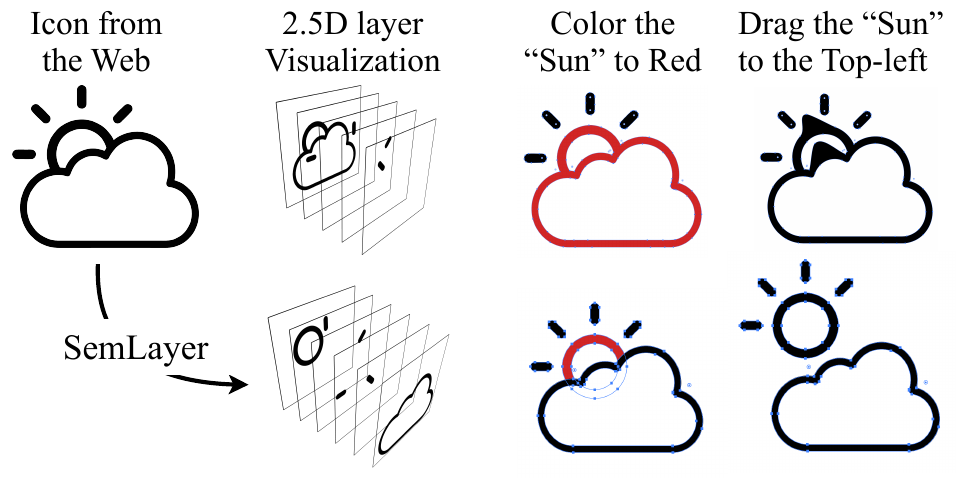}
    \caption{\ourmodel can convert web-sourced icon into designer-friendly, semantically layered representations, enabling intuitive editing (bottom row) that would otherwise be difficult (top row).}
    \label{fig:teaser}
\end{figure}

Vector graphics provide resolution-independent representations built from geometric primitives, making them essential in digital design.
In contemporary workflows, vector assets are typically organized into multiple editable layers, with semantically meaningful primitives separated to support downstream manipulation. In practice, however, icons are often distributed in a \emph{flattened} form, where these layers are merged into a single compound path.
Such flattening commonly arises during export for third-party redistribution or through optimization for file size and rendering performance.
Once this semantic structure is lost, routine editing tasks (\eg, recoloring, restyling, animation, and part-level modification) become difficult, as illustrated in \cref{fig:teaser}. As a result, designers have to manually re-segment and reconstruct the artwork before further editing.

To address this challenge, we introduce the task of \emph{semantic layer construction} for abstract icons. The goal is to recover a layered, semantically meaningful representation from a single-path or compound-path input. Unlike standard vector segmentation or image decomposition, this problem involves reasoning about both visible and occluded geometry, as well as inferring the correct stacking order and occlusion relationships between parts.

This task is particularly challenging (see \cref{sec:problem_formulation}).
Abstract icons are highly simplified, often omitting realistic cues such as texture, shading, and depth, which can make traditional visual understanding methods struggle.
Many icons contain overlapping or occluded elements that must be both completed and layered in a semantically consistent way.
The resulting vector reconstruction must be compact and geometrically clean for storage efficiency and editability.
Additionally, we face the lack-of-data challenge common in this area, specifically the absence of a large-scale corpus of vector icons annotated with part-level semantic layers, which makes training and evaluation difficult.

To tackle these challenges, we present \textbf{\ourmodel}, a visual generation-driven pipeline that restores editable layered structure from a flattened abstract icon.
Given an input monochrome image, our method first produces a segmentation that identifies distinct regions as potential semantic components.
Next, a semantic completion module reconstructs full object-level shapes, extending beyond visible boundaries to recover plausible occluded geometry.
Finally, we assemble the recovered parts into a coherent layered vector representation with inferred occlusion hierarchy, thereby completing the transition from raster image space to editable vector space.
Moreover, we introduce a scalable dataset construction workflow that leverages modern generative AI models to address this lack-of-data challenge, and curate \ourmodel-Segmentation, a diverse dataset of $8,567$ vector icons with segmented semantic parts, providing the first dataset for supervised learning and quantitative evaluation in semantic icon segmentation.

Through extensive qualitative and quantitative evaluations, we demonstrate that \ourmodel~produces semantically meaningful layers with significantly higher fidelity. Compared to the strongest baseline, our method improves segmentation by +$5.0$ mIoU and +$16.7$ PQ, and achieves the best completion performance with $85.2$ mIoU and a CD $46.6$. These gains translate directly into more accurate layer decomposition and more faithful reconstruction of occluded geometry, enabling new editing, animation, and design workflows for previously static vector graphics.

\vspace{-12pt}
\paragraph{Contributions}
Our main contributions are as follows:
\begin{itemize}
    \item We propose a generative pipeline that reformulates semantic segmentation as a controllable colorization problem and leverages diffusion-based priors for both segmentation and amodal completion, solving a task that no single existing model can address.
    \item We contribute two purpose-built datasets, \ourmodel-Segmentation and \ourmodel-Completion, the first large-scale resources for SVG semantic segmentation and amodal completion, together with a generation-based workflow that enables future research in this domain.
    \item We provide comprehensive evaluations demonstrating the practical benefits of semantic reconstruction for real-world design workflows, such as vector object recognition, editing, and animation.
\end{itemize}

%% file: sec/2_related.tex
\section{Related Works}
\label{sec:related}

\subsection{Layer-wise Image Vectorization}
\label{sec:related_image_vectorization}
Early image vectorization methods focus on developing explicit vector representations to faithfully reconstruct raster images, such as meshes~\cite{battiato2004svg, demaret2006image, swaminarayan2006rapid, lecot2006ardeco, liao2012subdivision, xia2009patch, yang2015effective, sun2007image} and curves~\cite{orzan2008diffusion, xie2014hierarchical, zhao2017inverse}. However, traditional approaches often overlook the \emph{layered structure} of vector art. In practice, artists create self-contained layers that are stacked to form the final composition. 
Motivated by this observation, recent work has shifted toward layer-wise vectorization, aiming to generate vector graphics in which each layer is complete to support downstream editing.
This setting poses two challenges: recovering occluded regions and inferring a plausible stacking order.
To address them, \emph{optimization-based} geometric methods cast vectorization as energy minimization that trades reconstruction fidelity against structural regularity (\eg, layer count, boundary smoothness).
They infer layer order from perceptual cues (\eg, T-/X-junctions) and complete occlusions using geometric or variational models~\cite{favreau2017photo2clipart, entem2018structuring, du2023image, law2025image}.
However, lacking semantic priors, these approaches can oversimplify hidden regions.
In parallel, differentiable rendering methods iteratively optimize primitives by minimizing reconstruction loss through a differentiable rasterizer~\cite{Li:2020:DVG}. Although visually faithful, they often produce an excessive number of layers and fragmented shapes~\cite{ma2022towards, hirschorn2024optimize, zhou2024segmentation, wang2024layered}.
More recently, \emph{learning-based} approaches learn depth and shape priors from artist-created data to improve shape completion quality~\cite {reddy2021im2vec, shen2021clipgen, chen2023editable, thamizharasan2024vecfusion, song2025layertracer, wu2025layerpeeler}. Their effectiveness, however, is constrained by the scarcity of large-scale vector graphics datasets, which limits generalization across styles and domains.
Concurrent work on sketch editing and animation also relates to ours. SketchEdit~\cite{ijcai2024sketchedit} enables stroke-level editing but assumes a clean part-level decomposition as input (\ie, individual strokes are already separated), which precisely motivates the need for our segmentation stage. FlipSketch~\cite{bandyopadhyay2024flipsketch} applies holistic motion transformations, lacking the precision required for part-level manipulation in professional workflows.
In this work, we target layer-wise vectorization for abstract, non-photorealistic icons~\cite{dominici2020polyfit, hoshyari2018perception, kopf2011depixelizing}. We mitigate data scarcity by leveraging the rich priors embedded in pretrained image generative and inpainting models, and we formulate layer-order recovery as a combinatorial optimization problem, achieving faithful layer reconstruction.

\subsection{Image Segmentation}
\label{sec:related_image_segmentation}
Image segmentation plays an important role in image analysis. Classical convolutional architectures~\cite{long2015fully, ronneberger2015u}, transformer-based models~\cite{strudel2021segmenter, zhang2023uni, li2024transformer}, and diffusion-based designs~\cite{amit2021segdiff, wu2024medsegdiff} have all been explored for this task. Nevertheless, their generalization can degrade when training data are limited~\cite{zhang2021understanding}. Recently, Segment Anything Model (SAM)~\cite{kirillov2023segment, ravi2024sam} has demonstrated strong zero-shot performance across diverse domains, including medical imaging~\cite{ma2024segment, deng2025segment, mazurowski2023segment, wu2025medical}, remote sensing~\cite{ren2024segment, chen2024rsprompter}, motion segmentation~\cite{xie2024moving}, and camouflaged object detection~\cite{tang2023can}.
In our setting, directly applying (or finetuning) SAM to icons in order to separate semantically distinct regions proves unreliable. The resulting masks are often noisy or over-/under-fragmented. We attribute this to the highly abstract nature of icons that have minimal texture, shading, and color.
To overcome this mismatch, we reformulate segmentation as colorization. Starting from a monochrome icon, we finetune image generation models to synthesize a chromatically differentiated rendering in which distinct colors align with semantic components, and then convert these regions into masks. This effectively bridges semantic understanding with reliable separation for abstract icons.

\subsection{Amodal Completion}
\label{sec:related_amodal_completion}
Amodal completion seeks to recover the full geometry and appearance of objects from partial observations.
The problem has been extensively studied~\cite{ao2023image}, with models typically trained on synthetic or richly annotated datasets~\cite{hu2019sail, yan2019visualizing, reddy2022walt, zhou2021human, ehsani2018segan, overlaybench, dhamo2019object, zheng2021visiting, tudosiu2024mulan, liu2024object} that cover natural scenes such as indoor environments, people, vehicles, and everyday objects. While these resources provide amodal masks via human annotation or procedural occlusions, they target photorealistic images and seldom provide the exact ground truth appearance for the hidden regions.
Our setting, which focuses on monochrome icons, differs substantially. To support icon specific completion, we curate a dataset that combines real world vector graphics with synthetic compositions. A key distinction from prior work is that, in SVG, layers are complete by construction, so the occluded shapes are available as exact ground truth rather than heuristic approximations.
Training on this dataset enables our completion module to learn the characteristic amodal patterns of icons, producing style consistent hallucinations.

%% file: sec/3_method.tex
\begin{figure*}[th]
    \centering
    \includegraphics[width=0.95\textwidth]{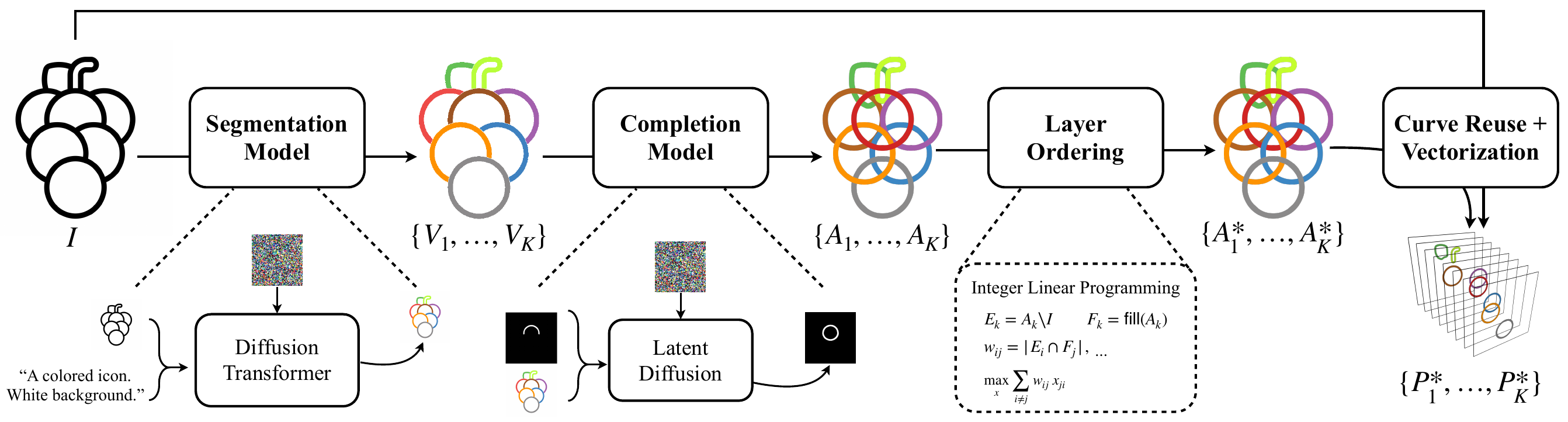}
    \caption{Overview of our \ourmodel{} pipeline.}
    \vspace{-12pt}
    \label{fig:method_overview}
\end{figure*}

\section{Method}
\label{sec:method}

Our method, \ourmodel, reconstructs a layered semantic representation from abstract icons, as illustrated in \cref{fig:method_overview}. We begin by formalizing the problem in \cref{sec:problem_formulation}, identifying four key challenges that arise when processing flattened icons. We then address these challenges through a three-stage pipeline. First, in \cref{sec:segmentation}, we introduce a semantic-aware generative segmentation approach that leverages diffusion models to decompose the icon into semantic parts. Second, in \cref{sec:completion}, we present an amodal layer completion method that recovers the full shape of each semantic part, including regions occluded by other parts. Third, in \cref{sec:ordering}, we propose a layer ordering strategy based on integer linear programming that determines the spatial arrangement of layers. The reconstructed and ordered layers are then vectorized back into the vector domain.

\input{sec/3_method/1_problem_formulation}

\input{sec/3_method/2_segmentation}

\input{sec/3_method/3_completion}

\input{sec/3_method/4_layering}

%% file: sec/3_method/1_problem_formulation.tex
\subsection{Problem Formulation}
\label{sec:problem_formulation}

Most contemporary icon platforms distribute abstract icons as flattened monochrome or duotone SVGs, typically represented as a single merged path or a collection of intricate compound paths. This flattening process introduces four primary challenges for icon decomposition:

\begin{enumerate}
    \item \textbf{Structural Ambiguity:} Multiple vector parameterizations can yield visually identical shapes, making it difficult to recover a unique structural representation from the flattened paths alone.
    
    \item \textbf{Semantic Degradation:} The abstraction process removes color cues and semantic annotations, making it challenging to identify and separate meaningful parts.
    
    \item \textbf{Hierarchical Fragmentation:} Flattening disrupts the original layer hierarchy by merging or truncating semantic components, resulting in incomplete part boundaries.
    
    \item \textbf{Occlusion-Order Indeterminacy:} Even after recovering the completed shape of each part, the original drawing order remains unknown.
\end{enumerate}

Let the input icon be represented as a set of Bézier paths $ P = \{P_1, \dots, P_N\} $ that define the visible silhouette. 

For the first challenge, to avoid ambiguities from vector parameterization, we rasterize the icon into a binary silhouette $ I \in \{0,1\}^{H \times W} $ for processing.

For the second and third challenges, we aim to decompose the silhouette into a set of visible semantic masks
\begin{equation}
V = \{V_1, \dots, V_K\}, 
\quad 
I = \bigcup_k V_k,
\end{equation}
and subsequently recover their completed amodal shapes
\begin{equation}
A = \{A_1, \dots, A_K\},
\quad 
V_k \subseteq A_k,
\end{equation}
where each amodal mask $A_k$ represents the full extent of a semantic part, including regions that occluded by others.

For the fourth challenge, to pursue a plausible layering order, we seek a permutation that arranges the set of unordered amodal masks into a properly ordered stack:
\begin{equation}
\{A_1, \dots, A_K\}
\rightarrow
\{A_1^*, \dots, A_K^*\},
\end{equation}
where the superscript $^*$ denotes the ordered sequence.

Finally, we vectorize the ordered layers back into the vector domain, producing fully editable, structurally coherent, and semantically organized icon components:
\begin{equation}
\{A_1^*, \dots, A_K^*\}
\xrightarrow{\mathrm{vec}}
\{\hat{P}_1, \dots, \hat{P}_K\}.
\end{equation}


Importantly, in the final vectorization stage, we do not generate curves from scratch. Instead, we adopt a \emph{curve reuse} strategy that maximally preserves the original high-quality Bézier segments and only repairs missing regions with newly constructed bridge curves (see \cref{sec:suppl_curve_reuse}).

%% file: sec/3_method/2_segmentation.tex
\subsection{Semantic-aware Generative Segmentation}
\label{sec:segmentation}

Semantic part segmentation of abstract icons is challenging due to high abstraction and minimal color cues. Classical segmentation methods like SAM~\cite{kirillov2023segment, ravi2024sam} often fail under such conditions, confusing strokes with filled regions and producing unstable masks. To address this, we leverage generative models that incorporate strong shape priors, offering more robust shape-semantic associations. We refer to this approach as \textit{Semantic-aware Generative Segmentation}.

We formulate segmentation as a colorization task: given a monochrome or duotone input, the model generates a colorized rendering where each distinct color corresponds to a semantic part. To preserve structural integrity while enabling semantic colorization, we adopt EasyControl~\cite{zhang2025easycontrol} as the implementation baseline, which enforces explicit structural conditioning in diffusion transformers~\cite{peebles2023dit, flux2024, labs2025flux1kontextflowmatching}. 

\paragraph{Training.}
We train on triplets $(p,\; I_{\text{cond}},\; I_{\text{tgt}})$ from our custom-built dataset (\cref{sec:trainset}), where $p$ is the text prompt for colorization, $I_{\text{cond}}$ is the binary silhouette $I$, and $I_{\text{tgt}}$ is a colorized icon with semantically separated parts. These inputs are then transformed into tokens $(z_p,\; z_c,\; x_0)$.

For a randomly sampled timestep $t \in [0,1]$, we draw noise $\varepsilon \sim \mathcal{N}(0, I)$ and construct the noisy latent
$
z_n = t\,\varepsilon + (1 - t)\, x_0.
$
We then concatenate it with the prompt tokens:
$$
z_{p\&n} = \operatorname{concat}[\, z_p,\; z_n \,].
$$

Following EasyControl~\cite{zhang2025easycontrol}, we utilize position-aware interpolation to align positional encodings with the generation resolution, while non-spatial conditions receive a positional-encoding offset to avoid overlap with the denoising tokens. The combined tokens are passed through a Transformer, where conditional LoRA~\cite{hu2022lora} modules are applied only to the condition-branch projections. Causal conditional attention
$
\mathrm{CrossAttn}(z_{p\&n},\; z_c)
$
enforces information flow from conditions to the denoising path.

We optimize a flow-matching objective~\cite{lipman2022flowmatching}:
$$
\mathcal{L}_{\mathrm{FM}}
=
\mathbb{E}_{t,\;\varepsilon \sim \mathcal{N}(0,I)}
\left\|
v_\theta(z_n,\; t,\; z_c)
-
(\varepsilon - x_0)
\right\|_2^{\,2},
$$
where $v_\theta$ predicts the velocity field that transports $z_n$ toward the clean latent $x_0$.

\paragraph{Inference.}
At test time, given a binary silhouette $I$, our model generates a colorized segmentation where each color represents a distinct semantic part:
\begin{equation}
I
\xrightarrow{\text{seg}}
\{\hat{V}_1, \dots, \hat{V}_K\}
\overset{\triangle}{=}
\hat{I},
\end{equation}
where $\hat{V}_i$ denotes the colorized version of the binary mask $V_i$, and $\hat{I}$ is the full colorized output. We extract the binary masks $\{V_1, \dots, V_K\}$ by thresholding each color channel, providing clean semantic part segmentations for subsequent amodal completion. 
Each $V_k$ corresponds to one \emph{semantic part instance} predicted by the model and is enforced to be a \emph{single connected component} in the image. Therefore, if a semantic object is split into multiple disjoint visible fragments due to occlusion or overlaps, the segmentation stage will output multiple instances $\{V_k\}$ (one per fragment), which are then handled independently in the subsequent amodal completion stage.

%% file: sec/3_method/3_completion.tex
\subsection{Amodal Layer Completion}
\label{sec:completion}

Given the semantic part segmentations from the previous stage, we next recover the complete amodal shape of each part. We build upon pix2gestalt~\cite{ozguroglu2024pix2gestalt}, a latent-diffusion-based amodal completion model originally designed for natural images. However, since such models are trained on photorealistic imagery, a substantial domain gap exists when applying them to vector-rendered icons. To address this, we curate a custom amodal-overlap dataset (\cref{sec:trainset}) and finetune the model on the icon domain.

\paragraph{Training.}
The model takes as input an occluded image $x_{\text{occ}}$ and its visible-region mask $m_{\text{vis}}$. Conditioning is provided to the latent-diffusion UNet through two streams: (1) a CLIP~\cite{radford2021clip} image embedding $c_{\text{clip}}=\mathrm{CLIP}(x_{\text{occ}})$ encoding high-level semantics, and (2) a concatenated latent input combining the VAE-encoded occluded patch $z_{\text{occ}}=E(x_{\text{occ}})$ and the visible-region mask. The UNet receives
\[
\tilde z_t = \mathrm{concat}\!\left(z_t,\; z_{\text{occ}},\; m_{\text{vis}}\right),
\]
and iteratively denoises from random noise while attending to $c_{\text{clip}}$, yielding the completed amodal shape $x_{\text{whole}}$.

We follow the standard noise-prediction objective. The ground-truth complete shape $x_{\text{whole}}$ is first encoded as
\[
z_0 = E(x_{\text{whole}}), \qquad
z_t = \sqrt{\bar\alpha_t}\, z_0 + \sqrt{1-\bar\alpha_t}\,\varepsilon,
\]
where noise $\varepsilon \sim \mathcal{N}(0, I)$ is added at timestep $t$. The UNet predicts the injected noise:
\[
\hat\varepsilon = \epsilon_\theta(\tilde z_t,\; t,\; c_{\text{clip}}).
\]
Then we minimize the mean-squared error:
\[
\mathcal{L} = \left\|\varepsilon - \hat\varepsilon\right\|_2^2.
\]

\paragraph{Training with Fragmented Visibility.}
In layered icons, it is common that a single semantic object is split into multiple disjoint visible fragments due to occlusion. To make the model robust to such cases, we explicitly simulate this setting during training: when an object is partially occluded and decomposed into multiple visible components $\{V^{(i)}\}$, we treat each visible fragment $V^{(i)}$ as an independent training input. Importantly, all such fragments share the same supervision target. This enforces a many-to-one completion behavior: regardless of which fragment is observed, the network is trained to recover the same underlying shape, which significantly improves robustness to heavy occlusions.

\paragraph{Inference.}
At test time, we apply the amodal completion model to each segmented visible part. The model outputs the completed amodal shape $A_k$:
\begin{equation}
\{V_1, \dots, V_K\},\; \hat{I}
\xrightarrow{}
\{A_1, \dots, A_K\}.
\end{equation}
The resulting amodal masks recover the full extent of each semantic part and guide our layer ordering.

\paragraph{IoU-based Completion Merging.}
Since multiple visible fragments from the same object are completed independently, the model may produce several highly similar amodal predictions. We therefore apply a merging step: for any pair of completed shapes $(A_i, A_j)$, if their intersection-over-union (IoU) exceeds a threshold $\tau$, we treat them as belonging to the same object instance and merge them into a single amodal layer. In practice, we use $\tau = 0.7$.

%% file: sec/3_method/4_layering.tex
\subsection{Layer Ordering Optimization}
\label{sec:ordering}

Given the completed amodal masks \(A = \{A_1, \dots, A_K\}\), we aim to determine a plausible layering order. While each amodal mask \(A_k\) represents the complete shape of a part, only a subset \(V_k \subseteq A_k\) is visible in the original silhouette \(I\). The completion process recovers extra pixels
\[
E_k = A_k \setminus I,
\]
which represent regions that should be occluded by other parts in a correct layering.

\paragraph{Fill Regions.}
For each part \(k\), we define a fill region \(F_k\) as the solid region enclosed by the \emph{outermost contour} of its vectorized amodal shape \(A_k\). That is, regardless of internal holes or cut-outs, \(F_k\) is treated as a fully filled region capable of occluding any underlying pixels. This is to ensure a strict layer stack representation without cyclic interleaving.

\paragraph{Occlusion Reward and Visibility Penalty.}
A correct layering should satisfy two criteria:
(1) Occlusion consistency: each extra pixel in \(E_i\) is covered by \emph{at least one} part;
(2) Visibility consistency: each visible pixel in \(V_i\) is \emph{not} incorrectly covered by other parts. 
In both cases, each pixel is counted \emph{only once}: covering the same pixel multiple times gives no additional reward, and occluding the same visible pixel multiple times incurs no additional penalty.

\paragraph{ILP Formulation.}
Let the unknown ordering be a permutation of \(\{1,\dots,K\}\). For each ordered pair \((i,j)\), we define
\[
x_{ij} =
\begin{cases}
1, & \text{if part } i \text{ is above part } j, \\
0, & \text{otherwise}.
\end{cases}
\]
We enforce:
\[
x_{ij} + x_{ji} = 1 \qquad \forall\, i \neq j,
\]
and the transitivity constraints for all distinct triples \((i,j,k)\):
\[
x_{ij} + x_{jk} + x_{ki} \le 2,
\]

\paragraph{Pixel-wise Coverage Variables.}
For each part \(i\), we introduce two additional binary variables:
\begin{itemize}
    \item \(y_i = 1\) if the extra region \(E_i\) is covered by \emph{at least one} part drawn above it.
    \item \(z_i = 1\) if the visible region \(V_i\) is incorrectly occluded by \emph{at least one} part drawn above it.
\end{itemize}

To link these variables to the ordering, we define the following precomputed binary indicators:
\[
c_{ij} =
\begin{cases}
1, & \text{if } |E_i \cap F_j| > 0, \\
0, & \text{otherwise},
\end{cases}
\qquad
d_{ij} =
\begin{cases}
1, & \text{if } |V_i \cap F_j| > 0, \\
0, & \text{otherwise}.
\end{cases}
\]

We then impose the logical constraints:
\[
y_i \le \sum_{j \neq i} c_{ij} \, x_{ji},
\qquad
z_i \le \sum_{j \neq i} d_{ij} \, x_{ji},
\]
which ensure that \(y_i\) can only be activated if at least one upper layer indeed covers the corresponding region.

\paragraph{Objective.}
Our final objective balances rewarding correct occlusion of extra regions and penalizing incorrect occlusion of visible regions:
\[
\max_{x,y,z}
\quad
\sum_i y_i \;-\; \lambda \sum_i z_i,
\]
where \(\lambda > 0\) controls the trade-off between occlusion consistency and visibility preservation. We set \(\lambda = 1\).

\paragraph{Solution.}
Once the ILP is solved and the optimal binary assignment \(x^*\) is obtained, we extract the corresponding permutation \(\pi^*\), produce the ordered layers:
\[
x^* \rightarrow \pi^* \rightarrow \{A_1^*, \dots, A_K^*\}.
\]
This ordered stack is then vectorized to obtain the final editable icon representation.

%% file: sec/4_dataset.tex
\begin{table*}[t]
\centering
\small
\setlength{\tabcolsep}{0.75em}
\begin{tabular}{l|cc|cc|cc}
\toprule
\multirow{2}{*}{\textbf{Segmentation Model}} &
\multicolumn{4}{c|}{\textbf{Segmentation}} &
\multicolumn{2}{c}{\textbf{Completion}} \\
\cmidrule(lr){2-5}\cmidrule(lr){6-7}
& mIoU (\%) $\uparrow$ & PQ (\%) $\uparrow$ & mIoU$_{\text{Refined}}$ (\%) $\uparrow$ & PQ$_{\text{Refined}}$ (\%) $\uparrow$
& mIoU (\%) $\uparrow$ & CD $\downarrow$ \\
\midrule
gpt-image-1~\cite{openai_gpt_image_1}
& $25.4$ & $6.20$ & $57.2$ & $39.3$ & $60.9$ & $71.4$ \\
SAM2~\cite{ravi2024sam}
& $51.1$ & $26.2$ & $62.2$ & $37.8$ & $69.2$ & $61.7$ \\
SAM2$^{*}$
& $79.3$ & $59.4$ & $85.3$ & $78.0$ & $80.7$ & $49.1$ \\
Ours
& $\mathbf{84.3}$ & $\mathbf{76.1}$ & $\mathbf{86.4}$ & $\mathbf{78.3}$ & $\mathbf{85.2}$ & $\mathbf{46.6}$ \\
\bottomrule
\end{tabular}
\caption{
Comparison of different segmentation models.
Segmentation results are reported for both the \textit{Original} predictions and the \textit{Refined} predictions, where refinement assigns each unlabeled (black) pixel the color of its nearest labeled pixel.
}
\vspace{-12pt}
\label{tab:two_stage_results}
\end{table*}

\section{Dataset}

\subsection{Training Set}
\label{sec:trainset}

\subsubsection{\ourmodel-Segmentation}
\label{sec:data_segmentation}

We build this dataset from two sources: (1) real-world SVGs from LayerPeeler~\cite{wu2025layerpeeler}, and (2) a synthetic set generated using GPT-4o and gpt-image-1. For the real-world portion, we filter icons to contain $4$ to $10$ paths, each with exactly one \texttt{Z} command to ensure a single closed contour. Icons are abstracted by removing fills and rendering black strokes, followed by manual quality verification. This yields $4,920$ curated icons. For the synthetic portion, we obtain color-agnostic structural descriptions from GPT-4o and synthesize stroke-based icons via gpt-image-1. After human inspection, $3,647$ icons are retained. In total, the proposed dataset contains $8,567$ training data.

\subsubsection{\ourmodel-Completion}
\label{sec:data_completion}

We also construct the overlap dataset for completion using the LayerPeeler~\cite{wu2025layerpeeler} collection. For each sample, two distinct icons are selected as the object and occluder. The occluder is resized and positioned to create meaningful occlusions. Parallelized rejection sampling ensures quality and diversity. Each sample provides the occluded composite image, the full occluded object, and a binary visible-region mask. In total, \ourmodel-Completion dataset contains $50,000$ training triplets.

\subsection{Test Set}

For evaluation, we obtain an additional set of $48$ high-quality real-world SVG icons using the same filtering and abstraction pipeline as in \cref{sec:data_segmentation}, while ensuring no overlap with any training samples. These icons exhibit rich semantic structures and meaningful part-level occlusions. All segmentation and amodal completion results reported in this paper are evaluated on this set of $48$ real-world icons.

%% file: sec/5_experiment.tex
\section{Experiment}
\label{sec:exp}

\subsection{Implementation Details}

The segmentation model is trained from scratch for $40,000$ steps (lr is $1\times10^{-4}$, CFG scale is $4.5$) on the dataset in \cref{sec:data_segmentation}, and uses 25 sampling steps at $512\times512$ resolution during inference.
A nearest-neighbor refinement assigns every unlabeled (black) boundary pixel to its closest labeled color, yielding crisp, fully assigned regions.
The completion model is fine-tuned for $50,000$ steps (lr is $1\times10^{-5}$) on the dataset in \cref{sec:data_completion}, with 50 sampling steps at $256\times256$.
A lightweight post-processing step retains the largest connected component and enforces topological consistency to remove small artifacts.
Vectorization is performed with \texttt{potrace}~\cite{selinger2003potrace}.
All experiments run on $8$ NVIDIA A100 GPUs.
See further details in \cref{sec:suppl_impl}.

\subsection{Evaluation Metrics}

To evaluate the semantic-aware generative segmentation model, we report mIoU and Panoptic Quality (PQ)~\cite{kirillov2019panoptic} on the visible segmented regions.
For the completion model, we evaluate two metrics. (1) For semantic parts that don't require completion, the model should preserve visible geometry without introducing extraneous modifications; we therefore compute mIoU with respect to the reference masks. (2) For semantic parts requiring completion, we quantify geometric deviation from the reference invisible regions using the Chamfer Distance (CD), computed by sampling 4{,}096 points from each segment mask. 

\subsection{Quantitative Results}

\subsubsection{Segmentation Model}

We evaluate the segmentation model across two sequential stages: Segmentation and Completion.  
Stage~1 measures the performance of the generative segmentation model.  
Stage~2 measures the quality of completion results produced by a fixed completion module applied to the outputs of Stage~1.  
Thus, improvements at the completion stage indirectly indicate higher-quality segmentation inputs.

As shown in \cref{tab:two_stage_results}, our method surpasses gpt-image-1~\cite{openai_gpt_image_1}, SAM2~\cite{ravi2024sam}, and a fully fine-tuned SAM2 (SAM2$^{*}$). We achieve the highest mIoU and PQ in both the original and refined segmentation results, indicating more accurate semantic predictions and fewer unlabeled regions. Due to the stochastic nature of generative models, all experiments of ours and gpt-image-1 are executed three times with different seeds, and the mean values are presented. As the completion module is fixed, variations in completion metrics stem from the segmentation stage. Our approach produces the best visible-region mIoU and the lowest CD, reflecting more faithful reconstruction of occluded geometry. 

\subsubsection{Completion Model}

\begin{table}[th]
\centering
\resizebox{\linewidth}{!}{
\setlength{\tabcolsep}{2em}
\begin{tabular}{l|c|c}
\toprule
\textbf{Completion Model} & mIoU (\%) $\uparrow$ & CD (pix) $\downarrow$ \\
\midrule
gpt-image-1 & 10.7 & 98.6 \\
MP3D\cite{zhan2024mp3d} & 70.5 & 79.4 \\
MP3D-finetuned \cite{zhan2024mp3d} & 75.3 & 68.9 \\
Ours & \textbf{85.2} & \textbf{46.6} \\
\bottomrule
\end{tabular}
}
\caption{
Comparison of different completion models. 
}
\vspace{-18pt}
\label{tab:completion_models}
\end{table}

With the segmentation model fixed, we evaluate different completion models with identical segmentation inputs. As the models are generative and may exhibit randomness across runs, we run each method three times with different seeds and report the average results.

As shown in \cref{tab:completion_models}, \ourmodel offers substantially stronger completion performance. It best preserves visible geometry and most accurately predicts occluded structures.

\subsection{Qualitative Results}

\begin{figure}[th]
    \centering
    \vspace{-12pt}
    \includegraphics[width=\linewidth]{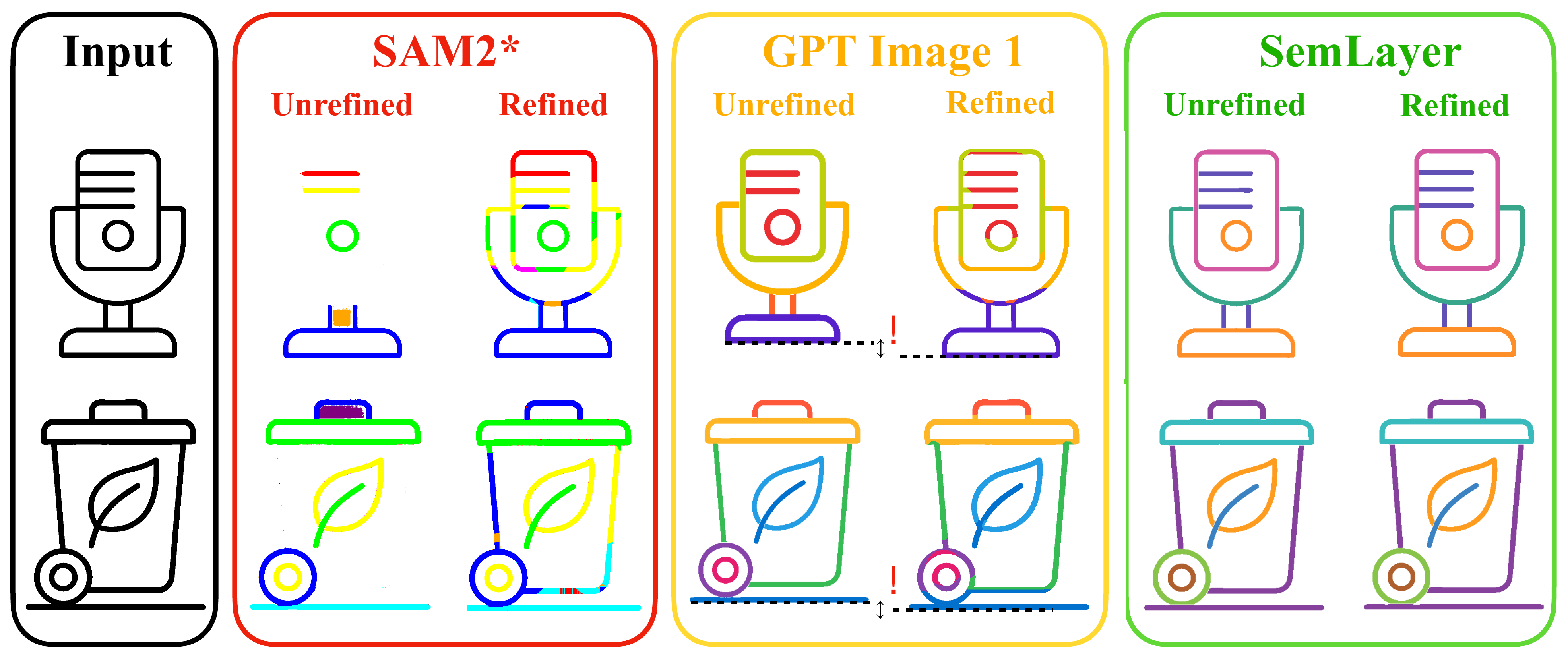}
    \caption{Qualitative comparison of segmentation quality.}
    \vspace{-12pt}
    \label{fig:qualitative_segmentation}
\end{figure}

First, we compare our segmentation model with existing segmentation methods, as shown in \cref{fig:qualitative_segmentation}. SAM2* frequently generates fragmented or improperly aligned color regions, whereas gpt-image-1 often has difficulty maintaining the identity and structural integrity of the original input. In contrast, our approach generates clean, complete, and structurally consistent segmentation that better preserves object boundaries and semantic regions. 

\begin{figure}[th]
    \centering
    \vspace{-12pt}
    \includegraphics[width=0.8\linewidth]{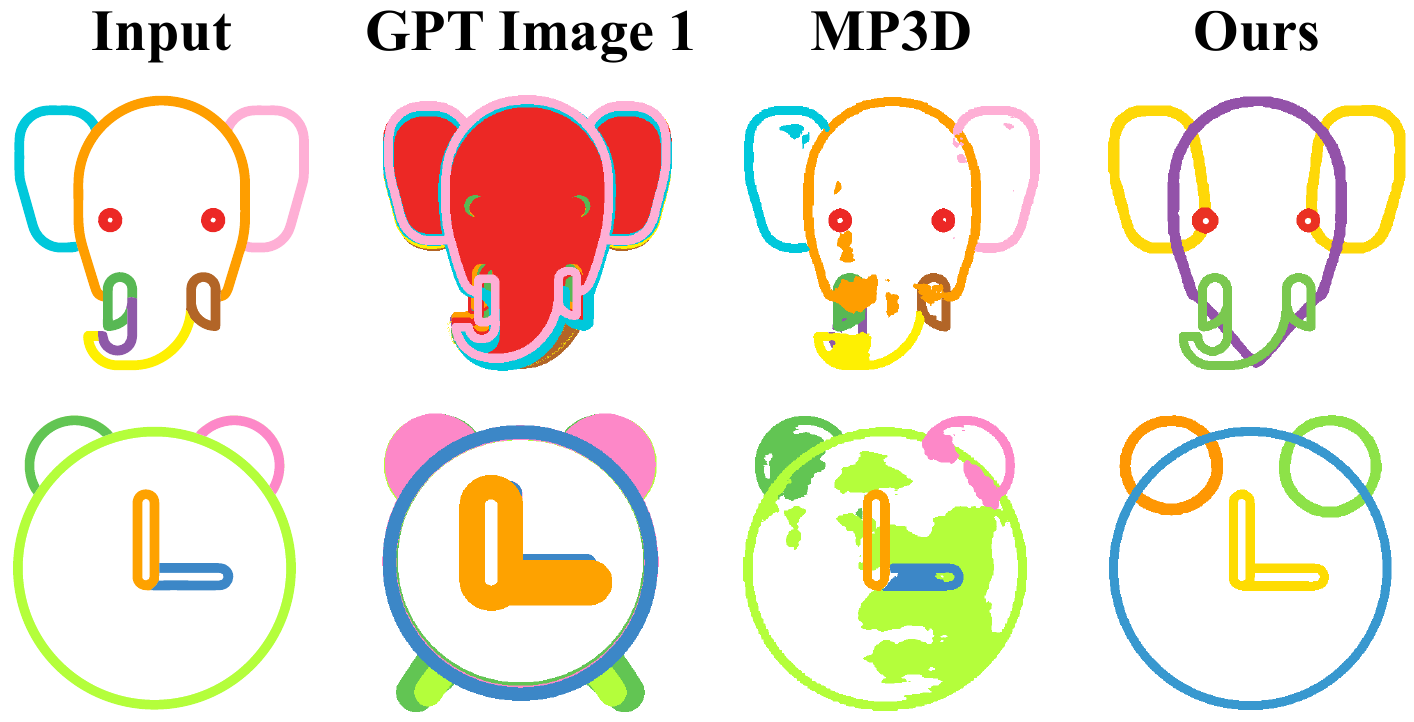}
    \caption{Qualitative comparison of completion results.}
    \vspace{-12pt}
    \label{fig:qualitative_completion} 
\end{figure}

We also compare our completion model with other completion methods, as shown in \cref{fig:qualitative_completion}. gpt-image-1 frequently introduces unnatural fill-in artifacts, and MP3D often yields incomplete or distorted shapes. Our method faithfully reconstructs both visible and occluded geometry, producing coherent and stylistically consistent outputs.

\begin{figure*}[th]
    \centering
    \includegraphics[width=0.82\linewidth]{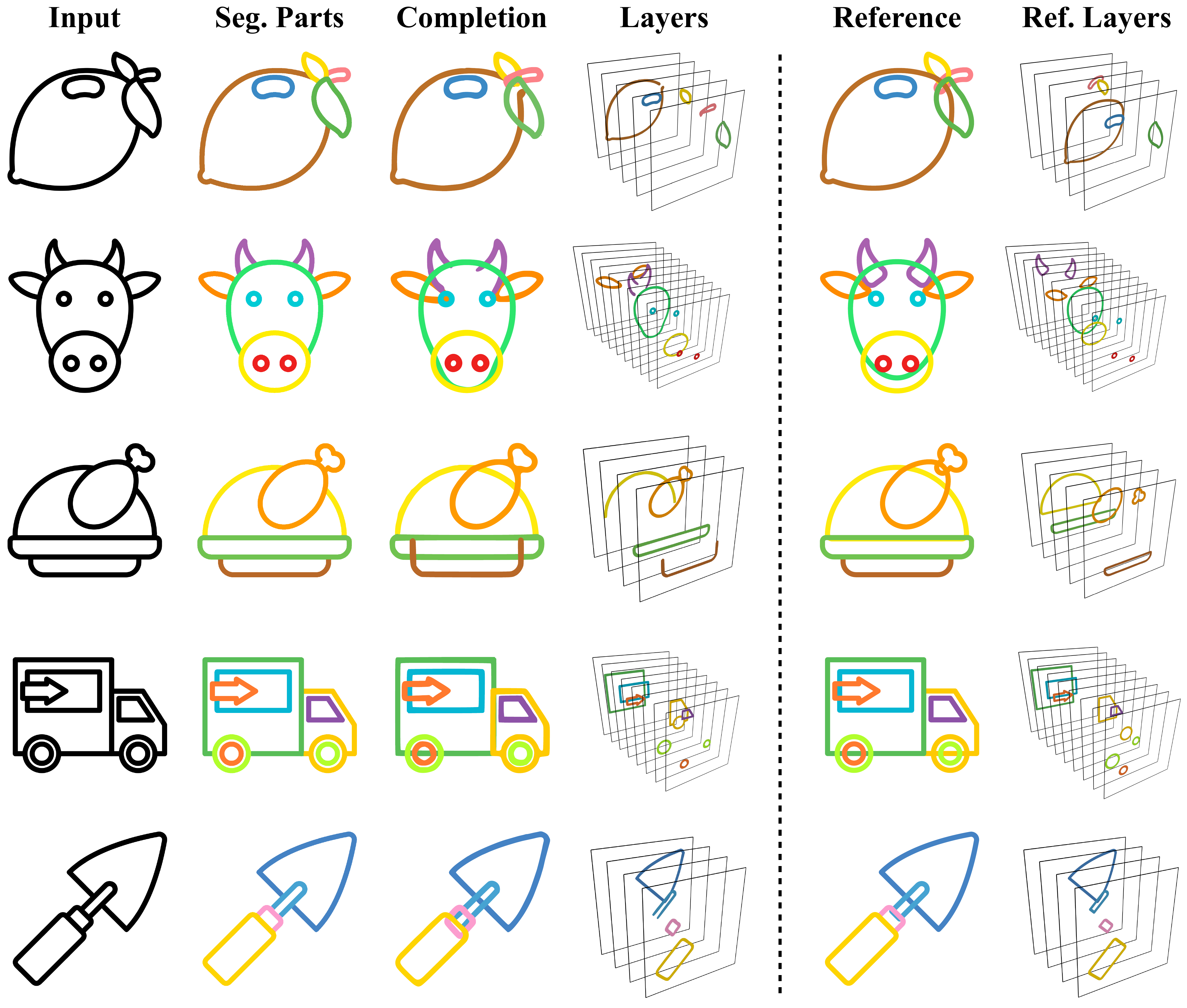}
    \caption{Progressive outputs from \ourmodel. The left four columns show: (1) the input icon, (2) results from our segmentation model, (3) the completed and vectorized output produced by the completion model with layer ordering, and (4) a 2.5D illustration visualizing separated layers. The right two columns present the reference layered SVG and its corresponding reference layer decomposition.}
    \label{fig:qualitative_pipeline}
\end{figure*}

Finally, we showcase several outputs of our \ourmodel, including intermediate results, as presented in \cref{fig:qualitative_pipeline}.

\subsection{Downstream Applications}
\vspace{-2pt}

\begin{figure*}[th]
    \centering
    \includegraphics[width=0.88\linewidth]{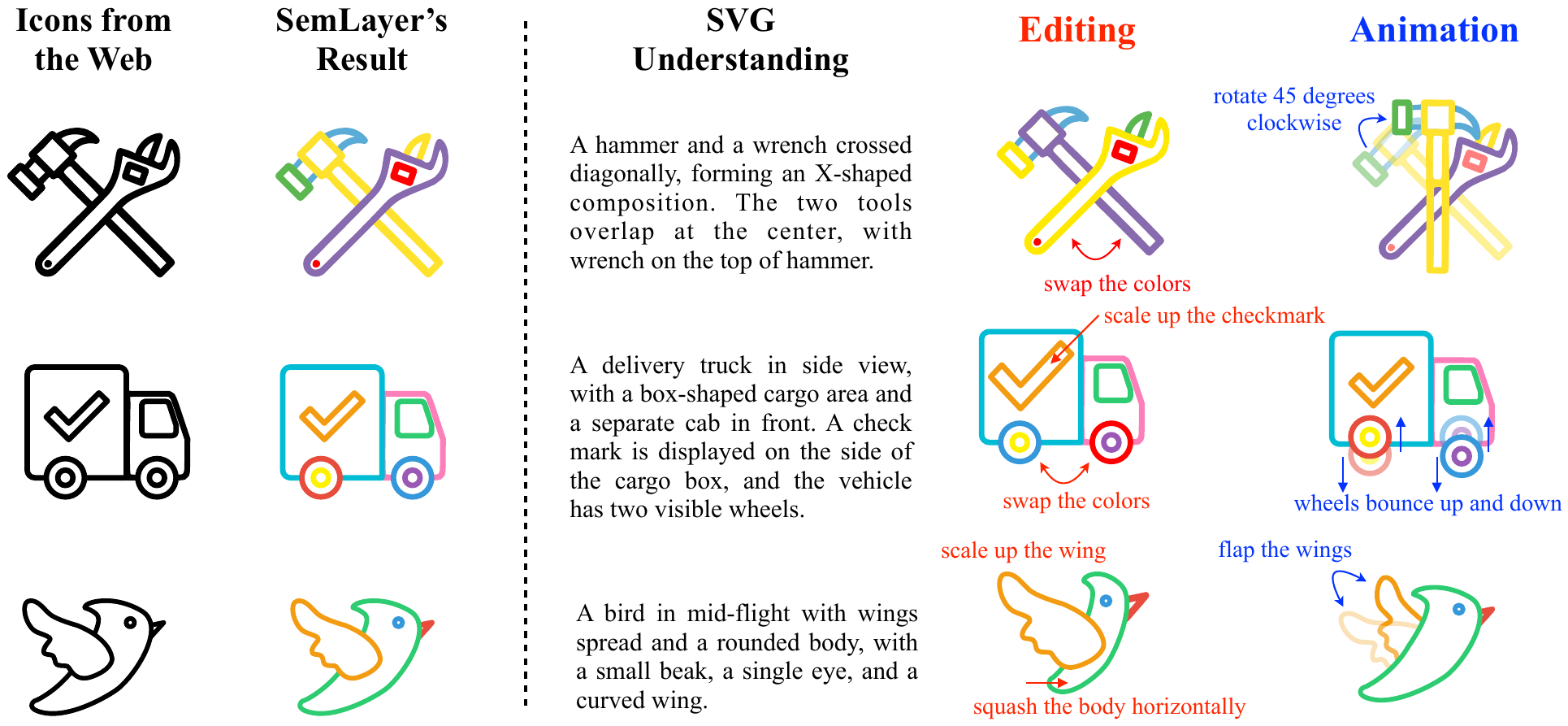}
    \caption{\ourmodel recovers a semantic-aware layered representation, enabling flexible part-level understanding, editing and animation that are difficult with the original icons, as shown in the figure.}
    \label{fig:application}
\end{figure*}

Recovering semantically structured layers enables applications that are difficult with flattened vector graphics. 
In typical web-distributed icons, semantic parts are merged into a single compound path, where components are represented as filled regions and internal holes. 
Editing such representations often leads to unintuitive or semantically incorrect results, as operations may only affect cut-out regions rather than meaningful object parts. 
In contrast, our semantic-layered representation separates these components into independent primitives, enabling local and semantically meaningful manipulation, as shown in \cref{fig:application}.

\noindent \textbf{SVG Semantic Understanding.}
Each recovered layer corresponds to a meaningful component, enabling part-level interpretation and reasoning. 
Icons can be decomposed into identifiable elements and spatial relations, which is more friendly for SVG semantic understanding.

\noindent \textbf{Part-level Editing.}
Semantic layers allow intuitive edits on individual components, such as recoloring, rotation, or rescaling, without reconstructing the icon structure. 

\noindent \textbf{Animation.}
Separated primitives enable simple animation through geometric transformations, such as rotating tools, bouncing wheels, or flapping wings. Achieving comparable effects using flattened icons is challenging.

%% file: sec/6_conclusion.tex
\vspace{-6pt}
\section{Conclusion}
\label{sec:conclusion}
\vspace{-2pt}

We addressed the challenge of restoring editable, semantically meaningful structure from flattened vector icons by introducing the task of semantic layer construction and proposing \ourmodel, a visual generation-based pipeline for segmenting, completing, and reassembling icon components into coherent layered vector graphics. Together with \ourmodel-Segmentation, the first dataset of icons annotated with semantic parts, our work establishes a foundation for systematic study and evaluation in this domain. 


Through extensive experiments, we show \ourmodel\ enables flexibility for rich editing and design workflows, which is impossible on flattened assets previously. 

\vspace{3pt}
\noindent \textbf{Limitations and Future Work.}
Our current pipeline focuses on black-and-white line drawings, which represent the most challenging setting for semantic decomposition. Since color itself is a strong semantic indicator, we expect the framework to extend to filled and colored icons given appropriate training data. Highly tangled or occluded icons can still challenge \ourmodel; we present failure cases in \cref{fig:failure_cases}. 
Looking ahead, we aim to extend \ourmodel to multi-color and stylistically diverse vector graphics, paving the way for fully automated vector editing tools that can adapt to a wide range of professional design needs.

\vspace{3pt}
\noindent \textbf{Ethics.}
\ourmodel is designed to assist designers to automate the decomposition of publicly distributed assets. It doesn't generate novel content and operates only on user inputs. Users should ensure they respect the input's license.

%% file: sec/X_suppl.tex

\clearpage
\setcounter{section}{0}
\renewcommand{\thesection}{\Alph{section}}
\maketitlesupplementary

\section{Implementation Details}
\label{sec:suppl_impl}

\subsection{Segmentation Post-processing}

Raw segmentation results can be influenced by anti-aliasing artifacts already present in the training data, which arise when the original segmentations are rasterized. Because supervision occurs on a pixel grid, boundaries between adjacent regions contain mixed or partially transparent pixels, making the interfaces inherently ambiguous. As a result, our predictions may show thin black gaps between segments, and boundary pixels may carry colors that are not strictly consistent with either side. To obtain clean, fully assigned regions, and to ensure positional consistency between the input mask and our prediction, since slight spatial shifts can occur—we treat all black pixels as unlabeled queries and assign each one the color of its nearest labeled pixel in the predicted map. This nearest-neighbor retrieval produces crisp, piecewise-constant regions with perfectly abutting boundaries, making them more robust to small geometric misalignments.

\subsection{Completion Post-processing}

Raw completion outputs may contain thin artifacts or small isolated noises near completion boundaries. We address these imperfections with a lightweight post-processing pipeline. First, we identify valid completion regions using a distance-based heuristic. We then retain the largest connected component, remove boundary rings via distance-transform filtering, and enforce topological consistency by preserving only regions connected to the expanded original mask. This yields clean, structurally coherent binary masks suitable for downstream vectorization.

\subsection{Curve Reuse}
\label{sec:suppl_curve_reuse}

A key advantage of working with parametric B\'ezier curves (instead of polygonal approximations) is that we can perform exact local edits while preserving the original high-quality geometry elsewhere. We formulate completion as a local curve surgery problem: maximally reuse the original curves and only replace the missing region with a small number of newly constructed bridge segments.

\begin{figure}[t]
    \centering
    \includegraphics[width=\linewidth]{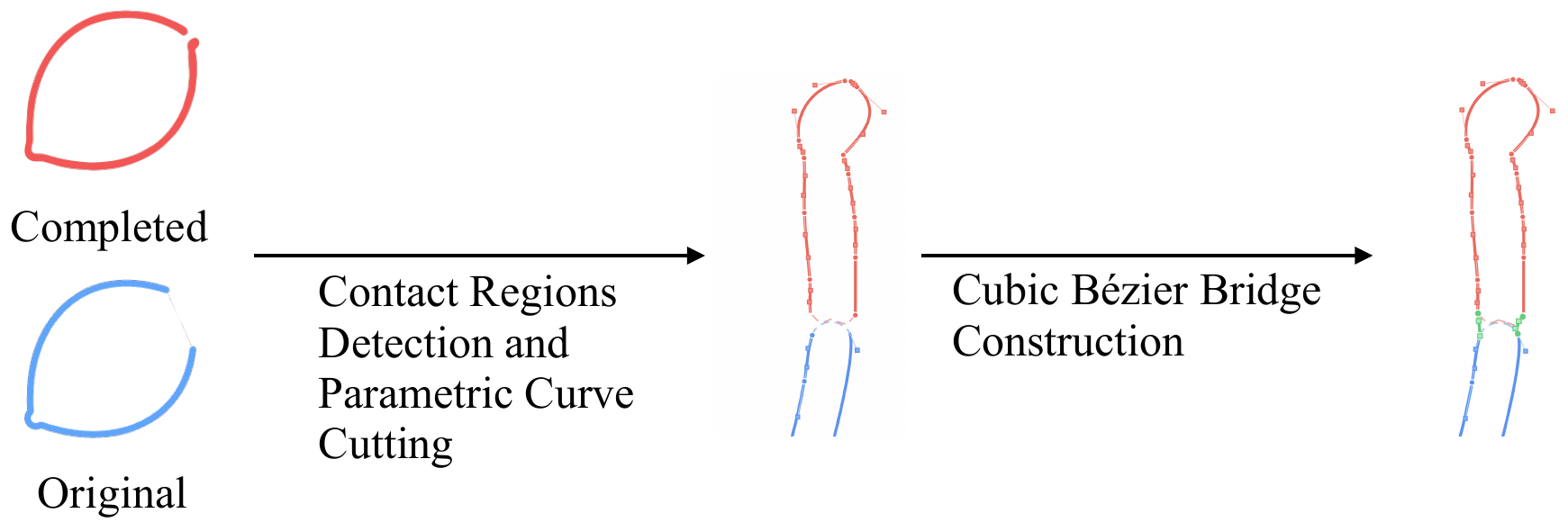}
    \caption{
    Illustration of our curve reuse and local surgery pipeline.
    We (1) detect the contact region between the original contour and the completed contour,
    (2) cut both curves locally,
    (3) keep the high-quality original segments (blue),
    (4) reuse the reliable part of the completion (red),
    and (5) connect them using newly constructed smooth bridge curves (green).
    }
    \label{fig:curve-surgery}
\end{figure}

Given an original incomplete contour $\mathcal{C}_A$ and a completed proposal $\mathcal{C}_B$, we first detect their contact regions by adaptive sampling and nearest-neighbor distance tests. These regions indicate where the two contours overlap and where replacement should happen. We then perform parametric curve cutting at the boundaries of each contact region. Unlike polygon-based methods, B\'ezier curves can be split exactly at arbitrary parameters, allowing us to remove only the local overlapping segments while keeping the rest of the curve bitwise identical to the original.

This produces two kept curve chains: (1) the preserved high-quality part of the original contour, and (2) the reliable part of the completion. To reconnect them, we construct cubic B\'ezier bridge curves that enforce $G^1$ continuity (matching position and tangent) at both ends. This ensures visually smooth transitions without kinks or cusps. Only these short bridge segments are newly created; all other geometry is reused verbatim. 

Finally, we assemble the merged contour by choosing the correct traversal order (forward or reversed) based on endpoint proximity.

\section{More Baselines}
\label{sec:suppl_ablation}

\subsection{Generative vs.\ Simpler Baselines.}
To validate the necessity of a generative approach, we implement classification and regression baselines adapted from the architecture of NIVEL~\cite{thamizharasan2024nivel}. As shown in \cref{tab:ablation_simpler}, these simpler models achieve substantially lower performance than our generative pipeline, confirming that the strong shape priors learned by diffusion models are crucial for both semantic segmentation and amodal completion of abstract icons.

\begin{table}[th]
\centering
\resizebox{\linewidth}{!}{
\setlength{\tabcolsep}{1.2em}
\begin{tabular}{l|c|c}
\toprule
\textbf{Method} & mIoU\textsubscript{Seg.}(\%) $\uparrow$ & mIoU\textsubscript{Comp.}(\%) $\uparrow$ \\
\midrule
Simpler Baseline & 46.6 & 63.9 \\
Ours & \textbf{84.3} & \textbf{85.2} \\
\bottomrule
\end{tabular}
}
\caption{Comparison with simpler classification/regression baselines. Our generative approach significantly outperforms non-generative alternatives.}
\label{tab:ablation_simpler}
\end{table}

\subsection{LLM-based Vector-Space Baselines}
\label{sec:suppl_llm_failure}

\begin{figure}[h]
    \centering
    \includegraphics[width=\linewidth]{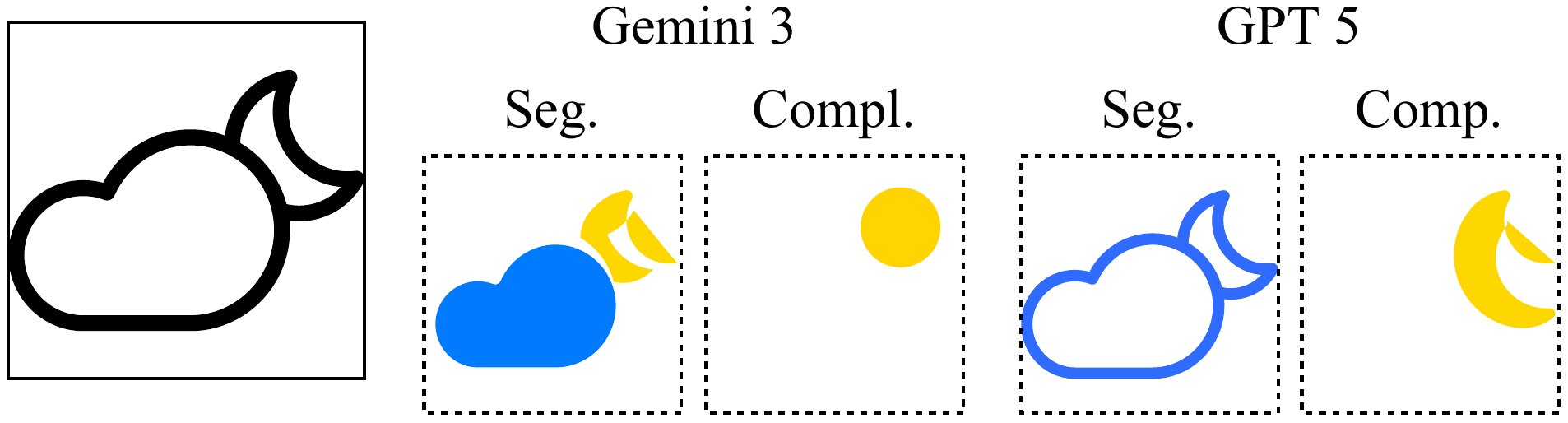}
    \caption{LLM-based vector-space baselines. Gemini-3 and GPT-5 fail to perform valid segmentation or completion when operating directly on SVG code.}
    \label{fig:llm_failure}
\end{figure}

A natural question is whether semantic layer construction can be performed directly in the vector domain by leveraging recent large language models (LLMs). To investigate this, we provided raw SVG code as input to Gemini-3 and GPT-5, prompting them to (1) assign semantic part labels to path groups (segmentation) and (2) complete occluded regions by generating new SVG paths (completion). As shown in \cref{fig:llm_failure}, both models fail to produce valid results: segmentation outputs are semantically incoherent, and completion outputs contain structurally broken or nonsensical paths. We attribute this to the fact that flattened SVG code consists of compound, semantic-less paths that lack the visual grounding necessary for spatial reasoning, validating our pixel-based approach.

\section{Runtime Analysis}
\label{sec:suppl_runtime}

\cref{tab:runtime} reports average per-icon runtime on a single NVIDIA A6000 GPU, evaluated over 48 test samples.
Our segmentation model runs on par with SAM2 and $3.7\times$ faster than gpt-image-1.
For completion, our model is $3\times$ faster than MP3D and over $22\times$ faster than gpt-image-1.

\begin{table}[th]
\centering
\resizebox{\linewidth}{!}{
\setlength{\tabcolsep}{0.8em}
\begin{tabular}{lc|lc}
\toprule
\multicolumn{2}{c|}{\textbf{Segmentation}} & \multicolumn{2}{c}{\textbf{Completion}} \\
Method & Sec/icon $\downarrow$ & Method & Sec/icon $\downarrow$ \\
\midrule
SAM2        & 14.32$\pm$3.72  & MP3D        & 49.27$\pm$17.21 \\
gpt-image-1 & 54.45$\pm$6.38  & gpt-image-1 & 379.73$\pm$153.96 \\
Ours        & \textbf{14.56$\pm$0.30}  & Ours        & \textbf{16.83$\pm$6.32} \\
\bottomrule
\end{tabular}
}
\caption{Runtime comparison. Average seconds per icon on a single NVIDIA A6000 GPU (48 test samples).}
\label{tab:runtime}
\end{table}

\section{Failure Cases}
\label{sec:suppl_failure}

\begin{figure}[h]
    \centering
    \includegraphics[width=0.5\linewidth]{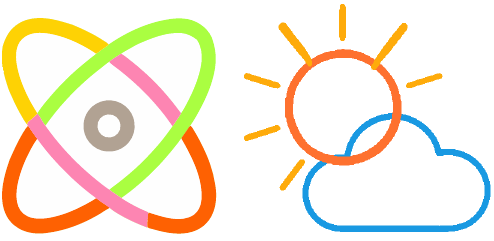}
    \caption{Representative failure cases. \textit{Left:} tightly tangled structures cause segmentation errors. \textit{Right:} the model fails to hallucinate entirely unseen structures (sun rays occluded by cloud).}
    \label{fig:failure_cases}
\end{figure}

We present representative failure cases in \cref{fig:failure_cases}. The first example illustrates failure under highly ambiguous and tightly tangled structures, where overlapping strokes make it difficult for the segmentation model to identify distinct semantic parts. The second example shows that our model struggles with pure generation: it fails to recover semantically expected yet entirely unseen structures (\eg, invisible sun rays fully occluded by overlapping elements). These cases highlight the remaining challenges in handling extreme occlusion and structural ambiguity.